\documentclass[runningheads]{llncs}

\usepackage[T1]{fontenc}
\usepackage{amsmath,amssymb,amsfonts}
\usepackage{algorithmic}
\usepackage{graphicx}
\usepackage{textcomp}
\usepackage{xcolor}
\usepackage{color, colortbl}
\definecolor{Gray}{gray}{0.9}
\definecolor{LightCyan}{rgb}{0.8,1,1}
\definecolor{Color2}{rgb}{1,1,0.8}
\newcommand{\R}{\mathbb{R}}

\begin{document}
\title{Large Language Models for Detection of Life-Threatening Texts}
%
%
\author{Thanh Thi Nguyen\thanks{Corresponding author.} \and
Campbell~Wilson \and
Janis~Dalins}

\authorrunning{T. T. Nguyen et al.}

\institute{AiLECS Lab, Faculty of IT, Monash University, Melbourne, VIC 3800, Australia
\email{\{thanh.nguyen9,campbell.wilson,janis.dalins\}@monash.edu}}

\maketitle       

\begin{abstract}
Detecting life-threatening language is essential for safeguarding individuals in distress, promoting mental health and well-being, and preventing potential harm and loss of life. This paper presents an effective approach to identifying life-threatening texts using large language models (LLMs) and compares them with traditional methods such as bag of words, word embedding, topic modeling, and Bidirectional Encoder Representations from Transformers. We fine-tune three open-source LLMs including Gemma, Mistral, and Llama-2 using their 7B parameter variants on different datasets, which are constructed with class balance, imbalance, and extreme imbalance scenarios. Experimental results demonstrate a strong performance of LLMs against traditional methods. More specifically, Mistral and Llama-2 models are top performers in both balanced and imbalanced data scenarios while Gemma is slightly behind. We employ the upsampling technique to deal with the imbalanced data scenarios and demonstrate that while this method benefits traditional approaches, it does not have as much impact on LLMs. This study demonstrates a great potential of LLMs for real-world life-threatening language detection problems. 

\end{abstract}
\section{Introduction}
\label{sec_int}
Identification of life-threatening language is crucial for several reasons. Detecting language indicating self-harm or harm to others allows for intervention before any harm occurs. This could involve alerting law enforcement, contacting emergency services, or providing support to individuals in crisis. Machine learning methods are frequently employed to identify harmful language like hate speech or offensive content. However, their application in detecting life-threatening language is relatively uncommon, especially in the English language. We found only a few such studies in the literature. For example, an approach was introduced in~\cite{mehmood2022threatening} for the automatic detection of threatening Urdu texts. Its performance was enhanced through the use of stacked classification models and feature extraction techniques. Likewise, the study in \cite{amjad2022overview1} proposed two shared tasks focused on detecting abusive and threatening language in Urdu. Both tasks are framed as binary classification challenges, where systems are tasked with categorizing Urdu tweets into two classes: abusive and non-abusive for the first task, and threatening and non-threatening for the second.

Recently, the work in \cite{rehan2023fine} devised a multilingual text classification framework to address the challenge of detecting threatening content across English and Urdu languages. However, similar to many traditional text classification methods, their approaches involve a text preprocessing phase including several steps such as \textit{removing} punctuations, hashtags, numbers, HTML tags, URLs, \textit{replacing} emoji and emoticons, \textit{addressing} the issue of misspelled words, and \textit{decoding} English abbreviations. Those preprocessing steps are \textit{tedious}, \textit{subjective}, and \textit{inconsistent} across datasets. This could significantly affect the performance and consistency of the classification models across different unseen test datasets. 

In this research, we propose an approach to life-threatening text detection using open-source large language models (LLMs) such as Gemma, Mistral, and Llama-2. By leveraging the power of LLMs, our approach is resistant to noisy data \textit{without text preprocessing steps} such as lemmatizing words, removing punctuation, hashtags, extra spaces, stop words, and so on, as seen in existing works. Our LLM-based approach \textit{avoids the need for separate designs} for feature extraction and classification, which otherwise might involve extensive trial and error to find the optimal combination of these steps. 
We show a consistent performance of the LLM-based methods across different data scenarios including balanced, imbalanced, and extremely imbalanced datasets. We provide a comprehensive comparison of our approach with traditional methods (TF-IDF, word embedding, topic modeling, and BERT) across six datasets and emphasize the importance of using not only accuracy, but also $F$-scores and AUC, especially when evaluating classification methods with class imbalance. 

\section{Competing Methods}
\label{sec_methods}

\subsection{TF-IDF}
TF-IDF is a method used to weight the terms in a bag-of-words model, which represents text data as a collection of words, ignoring grammar and word order. While TF-IDF provides a representation of text data in a vector space, it does not inherently acquire the semantic relationships between words like word embedding approaches.

\subsection{Word Embedding}
Methods like GloVe and Word2Vec learn semantic relationships by creating word embeddings based on co-occurrence patterns, where similar words have comparable representations.

\subsubsection{GloVe}
GloVe, short for Global Vectors for Word Representation, performs by creating a co-occurrence matrix, which contains the counts of how often each word appears within a context window of other words.

\subsubsection{Word2Vec}
This paradigm comprises two primary methods: continuous bag-of-words (CBOW) and skip-gram.
\textbf{CBOW} predicts the target word using its surrounding context words. \textbf{Skip-gram} works oppositely, predicting context words from a given target word.

\subsection{Topic Modeling}
This is an unsupervised learning technique used in NLP to discover latent topics or themes within a collection of documents. Common topic modeling approaches are Latent Dirichlet Allocation (LDA) and Latent Semantic Indexing (LSI).
\textbf{LDA} assumes each document is a mixture of topics, with each word assigned to one, iteratively adjusting topic distributions to maximize the likelihood of the given documents. \textbf{LSI} uses singular value decomposition on a term-document matrix for dimensionality reduction, uncovering latent semantic structures and building an index for efficient document retrieval.

\subsection{BERT}
BERT, representing bidirectional encoder representations from transformers, is a pretrained NLP model introduced in \cite{devlin2019bert}. Unlike previous language processing methods that handle text data sequentially in one direction (either left-to-right or right-to-left), BERT is bidirectional, meaning it can take into account the full context surrounding a word by examining both its left and right contexts concurrently. This helps obtain more precise representations of words and phrases. By fine-tuning BERT on specific text classification tasks, it can achieve great performance across various domains. In this study, we use the BERT-en-uncased model with its encoder and preprocessor respectively obtained from \cite{bert_encoder} and \cite{bert_preprocessor}.

\subsection{LLM - Gemma 7B}

Gemma was proposed by Google DeepMind, described in \cite{gemma_pdf}, and available in two variants: a 7 billion parameter model tailored for efficient development and deployment on GPU and TPU, alongside a 2 billion parameter model optimized for CPU and on-device applications \cite{gemma_pdf}. 

The 7B model employs multi-head attention, whereas the 2B model utilizes multi-query attention.
Gemma models utilize a subset of the SentencePiece tokenizer \cite{kudo2018sentencepiece} from Gemini. This tokenizer splits digits, retains extra whitespace, and employs byte-level encodings for unknown tokens. 
Instead of employing absolute positional embeddings, rotary positional embeddings \cite{su2024roformer} are utilized in every layer. Additionally, embeddings are shared between inputs and outputs to decrease model size \cite{gemma_pdf}.

\subsection{LLM - Mistral 7B}

Mistral 7B LLMs comprise the pretrained Mistral-7B-v0.1 model and several Mistral 7B Instruct variants \cite{jiang2023mistral}. These models are equipped with 7 billion parameters, which leverage grouped-query attention (GQA), and sliding window attention (SWA). 
GQA notably boosts inference speed and lowers memory needs during decoding, enabling larger batch sizes and thus increasing throughput. 

SWA is designed to handle extended sequences more efficiently while reducing computational costs, thus addressing a common constraint in using LLMs.
SWA utilizes the multiple layers of a transformer to extend its attention capability beyond a fixed window size $W$. Within each layer, the hidden state at position $i$, denoted as $h_i$, considers all hidden states from the preceding layer whose positions fall within the range of $i-W$ to $i$.

\subsection{LLM - Llama-2 7B}
The Llama-2 family, upon release, comprises pretrained and fine-tuned variants, each offering three models with 7B, 13B, and 70B trainable parameters. The pretrained variants were derived from a self-supervised learning method utilizing a vast text corpus containing two trillion tokens. Subsequently, these pretrained variants underwent further optimization via supervised fine-tuning procedure and reinforcement learning with human feedback, employing instruction datasets and human-annotated data tailored for dialogue use cases.

Instead of absolute positional encoding, Llama-2 uses rotary position embeddings to encode position information of tokens. For tokenization, the LLaMA tokenizer utilizes bytepair encoding sourced from SentencePiece \cite{kudo2018sentencepiece}. As a result, the Llama vocabulary envelops 32k tokens \cite{touvron2023llama}.

\section{Fine-tuning LLMs using LoRA}
\label{sec_fine-tuning_LoRA}

\begin{figure}[tb]
\centerline{\includegraphics[width=0.7\linewidth]{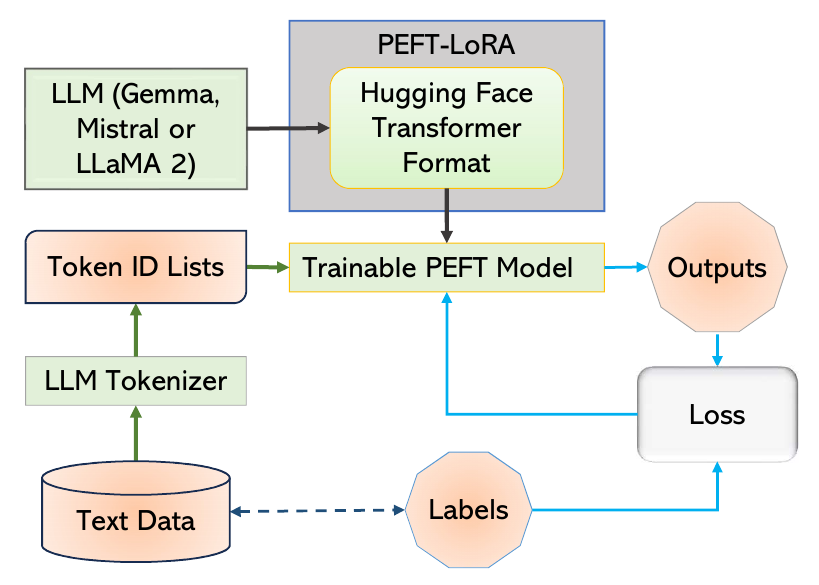}}
\caption{The method for fine-tuning a pretrained LLM for text classification tasks. Text data are first tokenized with an LLM tokenizer into token IDs, which are then used as inputs to the learnable PEFT model, constructed by transferring the original LLM weights to the Hugging Face format and loaded via the PEFT-LoRA mechanism.}
\label{fig_method}
\end{figure}

\begin{figure}[tbp]
\centerline{\includegraphics[width=0.95\linewidth]{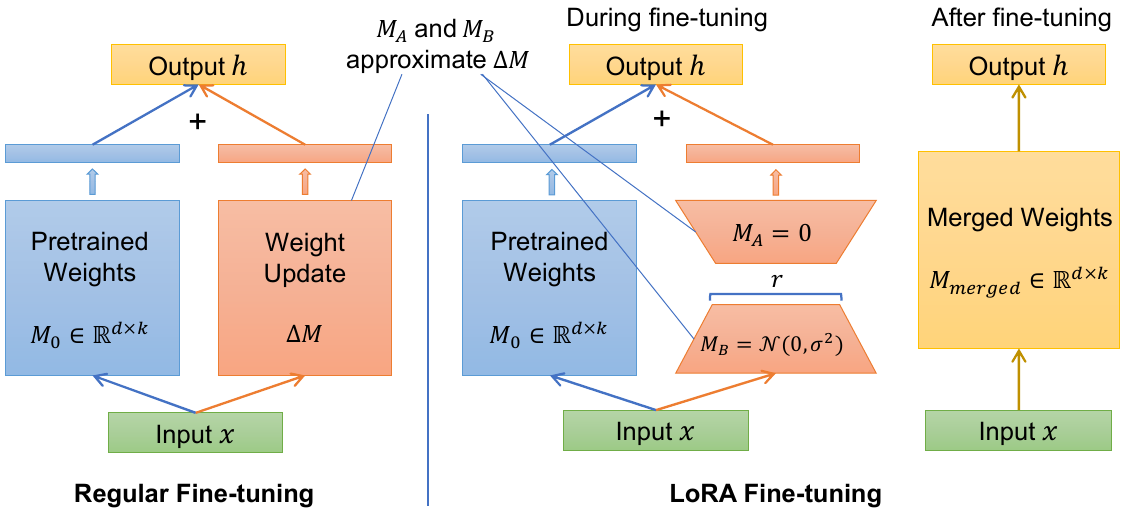}}
\caption{The difference between regular fine-tuning and LoRA fine-tuning. The LoRA matrices $M_A$ and $M_B$ approximate the weight update matrix $\Delta M$, with the inner dimension rank $r$ being a hyperparameter.}
\label{fig_lora}
\end{figure}

The LLMs such as Gemma, Mistral, and Llama-2 have amassed substantial world knowledge from pretraining on extensive text data. Our fine-tuning method, applied atop these models, will yield specialized expertise tailored to each fine-tuning dataset, while retaining the broad general knowledge and reasoning abilities acquired during the pretraining phase.

Our method for fine-tuning LLMs using low-rank adaptation (LoRA)~\cite{hu2021lora} is illustrated in Fig.~\ref{fig_method}. LoRA represents one of several parameter-efficient fine-tuning (PEFT) methods, including BitFit, SparseAdapter, AdaLoRA, among others, which can be utilized for fine-tuning LLMs \cite{lialin2023scaling}. The original weights of the LLMs are converted into the Hugging Face transformer format to leverage the fine-tuning tools provided by Hugging Face. These converted models are then integrated within the PEFT-LoRA framework, which employs LoRA based on the PEFT library \cite{peft}. 
The LoRA method performs low-rank fine-tuning by drawing inspiration from the structure-aware intrinsic dimension technique.
The update of parameters for a pretrained weight matrix $M_0 \in \R^{d \times k}$ is determined by the product of two low-rank matrices $M_A$ and $M_B$:
\begin{equation}
\Delta M = M_A M_B
\end{equation}
where $M_A \in \R^{d \times r}$ and $M_B \in \R^{r \times k}$ represent matrices of learnable parameters, with the rank $r$ being significantly smaller than the minimum of $d$ and $k$ \cite{lialin2023scaling}. The parameters of the pretrained model, $M_0$, remain fixed and do not undergo gradient updates during training. Both $M_0$ and $\Delta M = M_A M_B$ are applied to the same input $x$, as indicated by Hu et al. \cite{hu2021lora}; thus, the modification to the forward pass for $h=M_0 x$ is expressed as:
\begin{equation}
h = M_0 x + \Delta M x = M_0 x + M_A M_B x
\end{equation}
The matrix $M_A$ begins with zero initialization, while the matrix $M_B$ is initialized with random Gaussian values. Thus, $\Delta M = M_A M_B$ equals zero at the start of the fine-tuning process. Throughout training, $\Delta M x$ is scaled by a fixed factor $\frac{\alpha}{r}$, where $\alpha$ is a constant~\cite{hu2021lora}. Following training, the learnable parameters can be incorporated into the original weight matrix $M_0$ through the addition of the matrix $M_A M_B$. An illustration of LoRA and its difference to a regular fine-tuning method is presented in Fig. \ref{fig_lora}. LoRA enables the small trainable matrices $M_A$ and $M_B$ to undergo training, thereby adjusting to the new data, while simultaneously minimizing the total number of updates. By bypassing gradient computation for pretrained weights, the LoRA approach significantly reduces memory usage, enabling faster and more efficient fine-tuning.

Our fine-tuning methodology employs a cross-entropy loss function, comparing the output logits $\hat{y}$ of the neural network with the target $y$. In Fig. \ref{fig_method}, the term ``labels'' refers to the target $y$, while the ``outputs'' generated by the learnable PEFT model signify the logits $\hat{y}$. The cross-entropy loss function, averaged over the mini-batch size $N$, between the output $\hat{y}$ and the target $y$ is expressed as:
\begin{equation}
\mathcal{L}(\hat{y}, y) = \sum_{n=1}^N \frac{l_n}{\sum_{n=1}^N w_{y_n} }
\end{equation}
where $w$ denotes a weight vector with each element corresponding to a class, and $l_n$ is defined as:
\begin{equation}
l_n = - w_{y_n} \log \frac{\exp(\hat{y}_{n,y_n})}{\sum_{c=1}^C \exp(\hat{y}_{n,c})}
\end{equation}
where the logits $\hat{y}$ represent the unnormalized logits for each class, the target $y$ contains the class indices, and $C$ is the total number of classes.

The hyperparameter settings used in this study are as follows. The maximum length of a sentence is set to 128 tokens, either truncating longer sentences or padding shorter ones to ensure uniform size. The LoRA rank is set to 8, and the number of epochs is 10. We utilize the stochastic gradient descent AdamW optimizer for fine-tuning, with a learning rate of $2 \times 10^{-5}$ and an epsilon coefficient of $10^{-8}$. Additionally, the mini-batch size is fixed to 16 during training.

\section{Datasets Used}
\label{sec_datasets}
While identifying life-threatening language is important in law enforcement, healthcare, emergency services, mental health, and security agencies, collecting threatening texts for training a machine learning model is challenging due to its scarcity. 
Several hate speech datasets are available in the literature, yet the majority lack \textit{genuinely threatening} texts, except the ``dynamically generated hate speech'' dataset \cite{vidgen2021learning}. This dataset contains 18,969 not-hate texts and 22,175 hate texts. The hate texts are divided into several categories such as derogation, animosity, dehumanization and threatening where threatening texts take up 606 samples. We use these 606 samples as part of the threatening texts in our experiments. The other part of threatening texts are obtained from the Threatening English Language (TEL) corpus \cite{gales_2022_6815671}, which includes 309 written texts. In total, we have \textit{915 threatening texts} that comprise the data for the minor class.

The major class consists of non-threatening texts. To evaluate performance of the competing methods in dealing with balanced, imbalanced, and extremely imbalanced datasets, we create six scenarios for the non-threatening texts.

The first three scenarios are based on the 18,969 not-hate texts of the aforementioned ``dynamically generated hate speech'' dataset \cite{vidgen2021learning} where we sample without replacement 1,000, 5,000 and 10,000 samples from these not-hate texts. 

The other three scenarios for non-threatening texts are based on a public release of the dataset described in \cite{sachdeva2022measuring} that consists of 135,556 samples \cite{hugging_hate_speech_dataset}. The main outcome variable is the ``hate speech score'', which is a continuous hate speech measure, where a value less than 0.5 indicates a not-hate speech, and this part of data has 86,508 samples. We sample without replacement 1,000, 5,000 and 10,000 samples from these not-hate texts and consider them as non-threatening texts.

\begin{table}[tp]
\centering
\caption{The Compositions of the Six Experimental Datasets}
\label{table:datasets}
\begin{scriptsize}
\begin{tabular}{|c|c|l|c|}
\hline
 \textbf{Dataset} & \textbf{No. of Threatening} & \textbf{No. of Non-Threatening} & \textbf{Overall Characteristic} \\ 
\hline
1 & 915 texts in \cite{gales_2022_6815671} and \cite{vidgen2021learning} & \textbf{ 1,000} not-hate texts in \cite{vidgen2021learning} & Balance \\
\rowcolor{LightCyan}
2 & 915 texts in \cite{gales_2022_6815671} and \cite{vidgen2021learning} & \textbf{ 1,000} not-hate texts in \cite{hugging_hate_speech_dataset} & Balance \\ 
3 & 915 texts in \cite{gales_2022_6815671} and \cite{vidgen2021learning} & \textbf{ 5,000} not-hate texts in \cite{vidgen2021learning} & Imbalance \\
\rowcolor{LightCyan}
4 & 915 texts in \cite{gales_2022_6815671} and \cite{vidgen2021learning} & \textbf{ 5,000} not-hate texts in \cite{hugging_hate_speech_dataset} & Imbalance \\
5 & 915 texts in \cite{gales_2022_6815671} and \cite{vidgen2021learning} & \textbf{10,000} not-hate texts in \cite{vidgen2021learning} & Extreme Imbalance \\
\rowcolor{LightCyan}
6 & 915 texts in \cite{gales_2022_6815671} and \cite{vidgen2021learning} & \textbf{10,000} not-hate texts in \cite{hugging_hate_speech_dataset} & Extreme Imbalance \\ 
\hline
\end{tabular}
\end{scriptsize}
\end{table}

A summary of the six experimental datasets corresponding to the six aforementioned scenarios is presented in Table \ref{table:datasets}. Each dataset is randomly split into training (90\%) and testing (10\%) for experiments. To address the class imbalance issue, the upsampling technique is used by randomly sampling with replacement the threatening texts (i.e., the minor class) to match the number of non-threatening texts (i.e., the major class) in the training set.

\section{Performance Metrics}
\label{sec_metrics}
Each competing method is evaluated based on several metrics including accuracy, $F_\beta$ with $\beta$ equal to 0.5 (i.e., $F_{0.5}$-score), 1.0 ($F_1$-score), and 2.0 ($F_{2}$-score):
\begin{equation}
F_{\beta} = (1 + \beta^2) \cdot \frac{\text{P} \cdot \text{R}}{(\beta^2 \cdot \text{P}) + \text{R}}
\end{equation}
where $P$ is precision, $R$ is recall, and $\beta$ represents the relative importance of precision and recall. $F_{1}$-score determines the harmonic mean of precision and recall, striking a balance between the two metrics. It proves particularly useful in scenarios with imbalanced class distributions. $F_{0.5}$-score is a variant of $F_{1}$-score that weighs precision higher than recall. It is useful when precision is more important than recall. $F_{2}$-score is another variant of $F_{1}$-score that considers recall higher than precision.
We also employ area under the ROC curve (AUC) as it can quantify the ability of a machine learning model to distinguish between the positive and negative classes. In this study, we present all metrics as percentages.

\section{Results and Discussions}
\label{sec_exp_res}
Unlike our LLM-based approach, which functions as an end-to-end method, approaches using TF-IDF, word embedding, and topic modeling require integration with a classifier to classify textual features obtained from those techniques. For that purpose, we use an \emph{ensemble classifier} combining logistic regression, support vector machine, and random forest based on the \emph{soft voting} mechanism.

\subsection{Results on Balanced Datasets}

Results for the two balanced datasets, Dataset 1 and Dataset 2, are reported in \textbf{Table \ref{tab_dynamically_combined_dataset1}}.
TF-IDF achieves high scores in all metrics, while GloVe Embedding performs moderately, but with lower scores than TF-IDF. 
CBOW and skip-gram perform similarly, showing decent results but are outperformed by TF-IDF.
LSI outperforms LDA on all metrics in both datasets, but both have lower performance than TF-IDF and BERT.
Gemma is inferior to BERT, while Mistral exhibits excellent performance, surpassing most other methods in Dataset 2.
Llama-2 achieves the highest scores across all metrics in Dataset 1, but slightly lags behind Mistral in Dataset 2.

\begin{table}[ht]
\centering
\caption{Results on \textbf{Balanced} Datasets - Dataset 1 and Dataset 2} 
\begin{scriptsize}
\begin{tabular}{|l|c|c|c|c|c|c|c|c|c|c|}
\hline
\textbf{Competing}&\multicolumn{5}{c|}{\textbf{Dataset 1}} &\multicolumn{5}{c|}{\textbf{Dataset 2}} \\
\cline{2-11}
\textbf{Methods} & \textbf{\textit{ Acc. }} & \textbf{ \textit{$F_1$} } & \textbf{ \textit{$F_{0.5}$} } & \textbf{ \textit{$F_{2}$} } & \textbf{AUC} & \textbf{\textit{ Acc. }} & \textbf{ \textit{$F_1$} } & \textbf{ \textit{$F_{0.5}$} } & \textbf{ \textit{$F_{2}$} } & \textbf{AUC}\\
\hline
TF-IDF & 77.08 & 72.15 & 80.74 & 65.22 & 76.60 & 83.33 & 82.22 & 83.90 & 80.61 & 83.22 \\
\rowcolor{LightCyan}
GloVe Embedding & 69.27 & 59.31 & 71.43 & 50.71 & 68.57 & 74.48 & 70.66 & 75.84 & 66.14 & 74.14 \\
Word2Vec - CBOW & 71.88 & 68.24 & 72.32 & 64.59 & 71.59 & 69.79 & 69.15 & 68.71 & 69.59 & 69.79 \\
\rowcolor{LightCyan}
Word2Vec - Skip-gram & 71.35 & 68.57 & 71.26 & 66.08 & 71.15 & 70.83 & 68.18 & 70.59 & 65.93 & 70.64 \\
Topic Modeling - LDA & 67.19 & 62.72 & 66.75 & 59.15 & 66.88 & 71.35 & 68.57 & 71.26 & 66.08 & 71.15 \\
\rowcolor{LightCyan}
Topic Modeling - LSI & 72.40 & 69.36 & 72.64 & 66.37 & 72.16 & 80.73 & 79.10 & 81.59 & 76.75 & 80.56 \\
BERT - en-uncased & 79.17 & 77.53 & 79.68 & 75.49 & 79.02 & 83.33 & 80.00 & 88.64 & 72.89 & 82.89\\
\rowcolor{LightCyan}
LLM - Gemma - 7B & 75.52 & 74.59 & 74.84 & 74.35 & 75.48 & 80.73 & 78.11 & 83.12 & 73.66 & 80.43 \\
LLM - Mistral - 7B & 92.19 & 91.62 & 93.82 & 89.52 & 92.07 & \textbf{95.83} & \textbf{95.60} & 96.88 & \textbf{94.36} & \textbf{95.76} \\
\rowcolor{LightCyan}
\textbf{LLM - Llama-2 - 7B} & \textbf{93.23} & \textbf{92.82} & \textbf{94.38} & \textbf{91.30} & \textbf{93.14} & 95.31 & 94.97 & \textbf{97.25} & 92.79 & 95.19 \\
\hline
\end{tabular}
\label{tab_dynamically_combined_dataset1}
\end{scriptsize}
\end{table}

\subsection{Results on Imbalanced Datasets}
\subsubsection{Imbalanced Dataset 3 (Table \ref{tab_dynamically_combined_dataset4_5k})}
The competing methods are evaluated both with and without upsampling, where upsampling is used to mitigate the class imbalance issue.
Upsampling slightly improves TF-IDF and significantly enhances GloVe Embedding, CBOW, skip-gram, LDA, and LSI in terms of $F$-scores and AUC.
BERT and Gemma have comparable performance, while Mistral outperforms both in most metrics.
Llama-2 demonstrates excellent performance, but it is inferior to Mistral in all metrics. 
Upsampling does not lead to much improvement in accuracy, $F$-scores and AUC for Gemma, Mistral and Llama-2.

\begin{table}[ht]
\centering
\caption{Results on Dataset 3 - The First \textbf{Imbalanced} Dataset} 
\begin{scriptsize}
\begin{tabular}{|l|c|c|c|c|c|}
\hline
\textbf{Competing}&\multicolumn{5}{c|}{\textbf{Performance Metrics}} \\
\cline{2-6} 
\textbf{Methods} & \textbf{\textit{Acc.}} & \textbf{\textit{$F_1$}} & \textbf{\textit{$F_{0.5}$}} & \textbf{\textit{$F_{2}$}} & \textbf{AUC} \\
\hline
TF-IDF & 88.85 (89.02) & 56.00 (56.95) & 65.42 (66.15) & 48.95 (50.00) & 71.08 (71.62) \\
\rowcolor{LightCyan}
GloVe Embedding & 86.82 (88.18) & 32.76 (53.33) & 51.35 (62.31) & 24.05 (46.62) & 59.81 (69.80) \\
Word2Vec - CBOW & 84.63 (78.04) & 4.21 (38.68) & 9.90 (36.03) & 2.67 (41.75) & 51.08 (64.23) \\
\rowcolor{LightCyan}
Word2Vec - Skip-gram & 85.64 (83.28) & 17.48 (53.52) & 33.83 (49.74) & 11.78 (57.93) & 54.74 (74.33) \\
Topic Modeling - LDA & 85.14 (82.94) & 13.73 (51.67) & 27.13 (48.47) & 9.19 (55.33) & 53.56 (72.82) \\
\rowcolor{LightCyan}
Topic Modeling - LSI & 86.99 (82.60) & 39.37 (54.22) & 54.59 (49.11) & 30.79 (60.52) & 62.54 (75.68) \\
BERT - en\_uncased & 89.36 (78.38) & 59.35 (54.93) & 67.45 (45.51) & 53.00 (69.27) & 73.13 (80.61) \\
\rowcolor{LightCyan}
LLM - Gemma - 7B & 87.84 (85.64) & 56.63 (54.55) & 61.04 (54.37) & 52.81 (54.72) & 72.66 (73.11) \\
\textbf{LLM - Mistral - 7B} & \textbf{95.10} (92.06) & \textbf{83.80} (70.06) & \textbf{85.81} (78.80) & \textbf{81.88} (63.07) & \textbf{89.22} (78.67) \\
\rowcolor{LightCyan}
LLM - Llama-2 - 7B & 94.59 (94.09) & 81.40 (80.66) & 85.57 (82.02) & 77.61 (79.35) & 86.73 (87.74) \\
\hline
\multicolumn{5}{l}{$^{(*)}$\small Values in brackets correspond to cases where upsampling is used.}
\end{tabular}
\label{tab_dynamically_combined_dataset4_5k}
\end{scriptsize}
\end{table}

\begin{table}[ht]
\centering
\caption{Results on Dataset 4 - The Second \textbf{Imbalanced} Dataset} 
\begin{scriptsize}
\begin{tabular}{|l|c|c|c|c|c|}
\hline
\textbf{Competing}&\multicolumn{5}{c|}{\textbf{Performance Metrics}} \\
\cline{2-6} 
\textbf{Methods} & \textbf{\textit{Acc.}} & \textbf{\textit{$F_1$}} & \textbf{\textit{$F_{0.5}$}} & \textbf{\textit{$F_{2}$}} & \textbf{AUC} \\
\hline
TF-IDF & 92.91 (92.91) & 72.37 (73.08) & 83.59 (82.61) & 63.81 (65.52) & 79.17 (80.04) \\
\rowcolor{LightCyan}
GloVe Embedding & 88.18 (86.82) & 44.44 (55.68) & 62.22 (57.65) & 34.57 (53.85) & 64.55 (72.94) \\
Word2Vec - CBOW & 86.15 (80.91) & 25.45 (42.05) & 43.48 (40.92) & 17.99 (43.25) & 57.23 (65.93) \\
\rowcolor{LightCyan}
Word2Vec - Skip-gram & 87.50 (85.47) & 38.33 (56.12) & 57.21 (54.46) & 28.82 (57.89) & 61.96 (74.76) \\
Topic Modeling - LDA & 86.66 (83.78) & 32.48 (55.14) & 50.26 (51.13) & 23.99 (59.84) & 59.71 (75.51) \\
\rowcolor{LightCyan}
Topic Modeling - LSI & 90.71 (90.71) & 62.59 (72.08) & 74.43 (69.74) & 53.99 (74.58) & 73.93 (84.87) \\
BERT - en\_uncased & 91.55 (85.47) & 67.11 (63.56) & 77.51 (56.39) & 59.16 (72.82) & 76.62 (83.51) \\
\rowcolor{LightCyan}
LLM - Gemma - 7B & 91.55 (90.03) & 71.26 (63.80) & 74.34 (69.71) & 68.43 (58.82) & 81.43 (76.15) \\
\textbf{LLM - Mistral - 7B} & 95.78 (96.79) & 86.63 (89.02) & 86.35 (93.22) & \textbf{86.91} (85.18) & \textbf{92.25} (91.10) \\
\rowcolor{LightCyan}
\textbf{LLM - Llama-2 - 7B} & \textbf{96.96} (96.96) & \textbf{89.77} (89.66) & 92.94 (\textbf{93.53}) & 86.81 (86.09) & 92.07 (91.63) \\
\hline
\multicolumn{5}{l}{$^{(*)}$\small Values in brackets correspond to cases where upsampling is used.}
\end{tabular}
\label{tab_dynamically_combined_dataset5_5k}
\end{scriptsize}
\end{table}

\subsubsection{Imbalanced Dataset 4 (Table \ref{tab_dynamically_combined_dataset5_5k})}

Traditional methods with upsampling show improvements in $F$-scores and AUC compared to their non-upsampling counterparts. For example, the $F_1$-score of the GloVe Embedding method improves from 44.44\% to 55.68\%, while its AUC increases from 64.55\% to 72.94\%. On the contrary, \emph{upsampling does not significantly benefit LLMs}. Specifically, it only induces slight changes in the performance of Gemma, Mistral, and Llama-2.

Mistral and Llama-2 show consistent performance across multiple metrics, whether with or without upsampling. They obtain the highest overall accuracies, with Mistral achieving 95.78\% and Llama-2 reaching 96.96\%. Mistral attains the best AUC in this dataset, with a score of 92.25\%.

\subsection{Results on Extremely Imbalanced Datasets}
\subsubsection{Extremely Imbalanced Dataset 5 (Table \ref{tab_dynamically_combined_dataset4_10k})}

Due to the highly imbalanced nature of this dataset (and Dataset 6), the accuracy of all methods \emph{exceeds 91\%}, making $F$-scores and AUC \emph{more realistic metrics} for evaluations.

\begin{table}[ht]
\centering
\caption{Results on Dataset 5 - The First \textbf{Extremely Imbalanced} Dataset} 
\begin{scriptsize}
\begin{tabular}{|l|c|c|c|c|c|}
\hline
\textbf{Competing}&\multicolumn{5}{c|}{\textbf{Performance Metrics}} \\
\cline{2-6} 
\textbf{Methods} & \textbf{\textit{Acc.}} & \textbf{\textit{$F_1$}} & \textbf{\textit{$F_{0.5}$}} & \textbf{\textit{$F_{2}$}} & \textbf{AUC} \\
\hline
TF-IDF & 92.58 (92.58) & 31.93 (30.77) & 49.74 (49.18) & 23.51 (22.39) & 59.75 (59.27) \\
\rowcolor{LightCyan}
GloVe Embedding & 92.03 (91.39) & 21.62 (47.19) & 37.74 (49.18) & 15.15 (45.36) & 56.12 (70.05) \\
Word2Vec - CBOW & 91.48 (84.89) & 6.06 (33.73) & 13.51 (29.54) & 3.91 (39.33) & 51.53 (66.49) \\
\rowcolor{LightCyan}
Word2Vec - Skip-gram & 91.76 (89.10) & 15.09 (43.60) & 28.78 (41.14) & 10.23 (46.37) & 54.06 (70.70) \\
Topic Modeling - LDA & 91.94 (87.55) & 13.73 (33.33) & 28.46 (32.02) & 9.04 (34.76) & 53.68 (64.13) \\
\rowcolor{LightCyan}
Topic Modeling - LSI & 92.49 (88.55) & 28.07 (45.89) & 46.78 (41.47) & 20.05 (51.36) & 58.27 (73.73) \\
BERT - en\_uncased & 92.49 (87.45) & 25.45 (49.45) & 45.16 (41.93) & 17.72 (60.25) & 57.32 (79.80) \\
\rowcolor{LightCyan}
LLM - Gemma - 7B & 93.96 (93.22) & 57.14 (54.32) & 66.47 (60.61) & 50.11 (49.22) & 72.41 (72.00) \\
LLM - Mistral - 7B & 97.34 (96.15) & 83.80 (76.14) & 87.01 (79.95) & 80.82 (72.67) & 89.02 (84.56) \\
\rowcolor{LightCyan}
\textbf{LLM - Llama-2 - 7B} & \textbf{98.26} (97.99) & \textbf{89.62} (88.17) & \textbf{91.72} (89.32) & \textbf{87.61} (87.05) & \textbf{92.86} (92.71) \\
\hline
\multicolumn{5}{l}{$^{(*)}$\small Values in brackets correspond to cases where upsampling is used.}
\end{tabular}
\label{tab_dynamically_combined_dataset4_10k}
\end{scriptsize}
\end{table}

TF-IDF has better accuracy, $F$-scores, and AUC compared to GloVe Embedding. While upsampling does not improve TF-IDF, it significantly enhances GloVe Embedding in terms of $F$-scores and AUC. 

CBOW and skip-gram perform poorly, with extremely low $F$-scores. With upsampling, both their \emph{$F$-scores and AUC improve} significantly, while \emph{accuracy decreases}. 
LSI outperforms LDA in all metrics.
BERT exhibits equivalent performance to LSI, with both surpassing Word2Vec methods. Gemma's performance falls short of Mistral's. Llama-2 showcases outstanding performance, achieving the highest accuracy, $F$-scores, and AUC. Upsampling \emph{fails to enhance the performance} of LLM methods, i.e., Gemma, Mistral and Llama-2.

Irrespective of the metrics used, Llama-2 and Mistral emerge as top performers in this dataset.
Upsampling typically enhances or at least sustains the performance of traditional (non-LLM) methods. Nevertheless, even with upsampling, non-LLM methods consistently demonstrate inferior performance when compared with Mistral and Llama-2.

\subsubsection{Extremely Imbalanced Dataset 6 (Table \ref{tab_dynamically_combined_dataset6_10k})}

When using upsampling, there is a slight decrease in all metrics of TF-IDF.
Similar to Dataset 5, GloVe Embedding has lower accuracy, $F$-scores and AUC compared to TF-IDF. Upsampling \emph{improves its $F$-scores and AUC} significantly but \emph{not accuracy}. 

\begin{table}[ht]
\centering
\caption{Results on Dataset 6 - The Second \textbf{Extremely Imbalanced} Dataset}  
\begin{scriptsize}
\begin{tabular}{|l|c|c|c|c|c|}
\hline
\textbf{Competing}&\multicolumn{5}{c|}{\textbf{Performance Metrics}} \\
\cline{2-6} 
\textbf{Methods} & \textbf{\textit{Acc.}} & \textbf{\textit{$F_1$}} & \textbf{\textit{$F_{0.5}$}} & \textbf{\textit{$F_{2}$}} & \textbf{AUC} \\
\hline
TF-IDF & 94.60 (94.14) & 59.86 (52.94) & 72.61 (69.50) & 50.93 (42.76) & 72.76 (68.70) \\
\rowcolor{LightCyan}

GloVe Embedding & 92.77 (92.12) & 28.83 (52.22) & 50.31 (54.02) & 20.20 (50.54) & 58.42 (72.83) \\

Word2Vec - CBOW & 91.76 (87.91) & 13.46 (36.54) & 26.72 (34.73) & 9.00 (38.54) & 53.58 (66.24) \\
\rowcolor{LightCyan}

Word2Vec - Skip-gram & 92.12 (91.12) & 20.37 (52.68) & 37.41 (50.47) & 13.99 (55.10) & 55.69 (75.61) \\

Topic Modeling - LDA & 92.40 (92.31) & 27.83 (48.78) & 45.71 (53.91) & 20.00 (44.54) & 58.22 (69.60) \\
\rowcolor{LightCyan}

Topic Modeling - LSI & 93.32 (91.30) & 42.52 (56.62) & 60.54 (52.45) & 32.77 (61.51) & 63.96 (79.52) \\

BERT - en\_uncased & 93.77 (86.54) & 46.03 (51.16) & 66.21 (41.89) & 35.28 (65.70) & 65.16 (84.06) \\
\rowcolor{LightCyan}

LLM - Gemma - 7B & 97.16 (95.33) & 82.68 (67.52) & 85.85 (77.26) & 79.74 (59.95) & 88.45 (77.44) \\

LLM - Mistral - 7B & 98.26 (98.26) & 89.62 (89.50) & 91.72 (92.26) & 87.61 (86.91) & 92.86 (92.38) \\
\rowcolor{LightCyan}

\textbf{LLM - Llama-2 - 7B} & 98.08 (\textbf{98.44}) & 88.77 (\textbf{90.50}) & 89.63 (\textbf{93.97}) & \textbf{87.92} (87.28) & \textbf{93.23} (92.48) \\

\hline
\multicolumn{5}{l}{$^{(*)}$\small Values in brackets correspond to cases where upsampling is used.}
\end{tabular}
\label{tab_dynamically_combined_dataset6_10k}
\end{scriptsize}
\end{table}

BERT exhibits moderate performance, comparable to LSI, but worse than TF-IDF.
When employing upsampling, its accuracy and $F_{0.5}$ decrease, while $F_{1}$, $F_{2}$, and AUC show a significant increase.

Gemma performs exceptionally well, surpassing all traditional methods. However, when upsampling is applied, there is a slight decrease in all of Gemma's metrics.
Both Mistral and Llama-2 exhibit excellent performance, with very high accuracy, $F$-scores and AUC values, \emph{outperforming all other methods}. Upsampling does not have much impact on performance of these methods.

In summary, Mistral and Llama-2 consistently \emph{demonstrate superior performance} compared with traditional methods, indicating their effectiveness in handling text classification tasks, thanks to their sophisticated architectures. These LLM methods effectively \emph{address the class imbalance} issue, eliminating the need for upsampling, although upsampling does improve the performance of most traditional methods. Additionally, the results underscore the importance of \emph{$F$-scores and AUC} metrics, in addition to accuracy, when evaluating classification methods in the presence of class imbalance.

\section{Conclusions and Future Work}
\label{sec_con_fut}
This study explored various approaches, encompassing both traditional methods and LLMs, to differentiate between threatening and non-threatening texts. Results indicate that LLMs, in particular Mistral and Llama-2, consistently outperform traditional methods like word embedding, topic modeling, and BERT. To address class imbalance, we employed the upsampling technique on the training data. Upsampling generally enhances the capability of traditional methods in detecting minor class samples (i.e., threatening texts), resulting in improved $F$-scores and AUC values. However, even with upsampling, the performance of traditional methods still falls short compared to that of LLMs. Notably, LLMs exhibit remarkable efficacy in handling imbalanced data without requiring upsampling thanks to their sophisticated architectures.

Given the scarcity of threatening texts, particularly in low-resource languages, future endeavors could center on gathering such texts in those languages and exploring a multilingual approach based on LLMs. This multilingual model would seamlessly operate across languages without necessitating individual fine-tuning of LLMs on text data for each language. Another avenue for future research involves leveraging quantization methods to enhance the accessibility of LLMs. Different methods such as quantized low-rank adaptation, pruned and rank-increasing low-rank adaptation, and general pretrained transformer quantization can be explored as each has its own advantages and disadvantages.

\bibliographystyle{splncs04}
\bibliography{mybibliography}

\end{document}